\documentclass{article}

\usepackage{arxiv}

\usepackage[utf8]{inputenc} 
\usepackage[T1]{fontenc}    
\usepackage{hyperref}       
\usepackage{url}            
\usepackage{booktabs}       
\usepackage{amsfonts}       
\usepackage{nicefrac}       
\usepackage{microtype}      
\usepackage{lipsum,lineno,subcaption}
\usepackage{graphicx}
\usepackage{amsmath} 
\usepackage{natbib}
\usepackage{xcolor}
\setcitestyle{open={(},close={)}}

\usepackage{tabularray}
\UseTblrLibrary{booktabs}

\def\ie{{\em i.e.},\ }
\def\eg{{\em e.g.},\ }
\def\M{$M$}

\definecolor{black}{cmyk}{0, 0, 0, 1}
\definecolor{gray}{cmyk}{0, 0, 0, 0.25}
\definecolor{darkgray}{cmyk}{0, 0, 0, 0.5}

\DeclareMathOperator*{\argmin}{arg\,min}

\title{Modeling and Forecasting COVID-19 Cases using Latent Subpopulations}

\author{
 Roberto Vega \\
  Department of Computing Science\\
  University of Alberta\\
  Alberta Machine Intelligence Institute \\
  \texttt{rvega@ualberta.ca} \\
   \And
 Zehra Shah \\
  Department of Computing Science\\
  University of Alberta\\
  Alberta Machine Intelligence Institute \\
  \And
 Pouria Ramazi \\
  Department of Mathematics \& Statistics\\
  Brock University\\
  \And
 Russell Greiner \\
  Department of Computing Science\\
  University of Alberta\\
  Alberta Machine Intelligence Institute \\
  \texttt{rgreiner@ualberta.ca} \\  
}

\begin{document}

\maketitle
\begin{abstract}
Classical epidemiological models assume homogeneous populations. 
There have been important extensions to model {\em heterogeneous}\ populations, when the identity of the sub-populations is known, such as age group or geographical location. 
Here, we propose two 
new 
methods to model the number of people infected with COVID-19 over time, 
each as 
a linear combination of 
{\em latent}\ sub-populations --
\ie when we do not know which person is in which sub-population, and the only available observations are the 
{\em aggregates}\ across all sub-populations.
Method\#1 
is a dictionary-based approach, which begins with a large number of pre-defined sub-population models 
(each with its own starting time, shape, etc), then determines the (positive) weight of small 
(learned) number of sub-populations.
Method\#2 
is a mixture-of-\M\ fittable curves,
where \M, the number of sub-populations to use, is given by the user. 
Both methods are compatible with any parametric model; 
here we demonstrate 
their use with first (a)~Gaussian curves and then 
(b)~SIR trajectories. 
We empirically show the performance of the proposed methods, first in (i)~modeling the observed data and then in (ii)~forecasting the number of infected people 1- to 4-weeks in advance.
Across 187 countries, 
we show that the dictionary approach had the lowest mean absolute percentage error and 
also the lowest variance when compared with classical SIR models and
moreover, it was 
a strong baseline that outperforms many of the models developed for COVID-19 forecasting. 
\end{abstract}


\section{Introduction} 
Classical epidemiological models, such as the Susceptible-Infected-Removed model (SIR), have been extensively used to study how diseases spread. These models work under the assumption of a constant, homogeneous population, and so their performance degrades when analyzing heterogeneous populations --
\eg when the population consists of communities that have different behaviours, or that are geographically separated~\citep{calvetti2020metapopulation}.
In particular, the heterogeneity of disease transmission has a significant effect on compartmental model dynamics,
and ignoring this heterogeneity 
has has been shown to over-estimate
the height of the epidemic peak and the number of infected individuals at the end of the epidemic~\citep{dolbeault2020heterogeneous,dolbeault2021social}. 
Similarly, \cite{ellison2020implications} describes how using homogeneous models to describe an infectious disease in a heterogeneous population might lead to wrong conclusions in terms of disease forecasting, herd immunity, and the impact of government policies to control the spread of the disease. Hence it is crucial to develop models that consider population heterogeneity. This paper proposes two such methods that model the observed number of people infected by COVID-19 as a linear combination of independent, unknown, sub-populations. 
It also empirically demonstrates that these methods can effectively model, and forecast,
the number of infected people, across 187 countries.

\begin{figure}
\centerline{
 \includegraphics[width=0.7\textwidth]{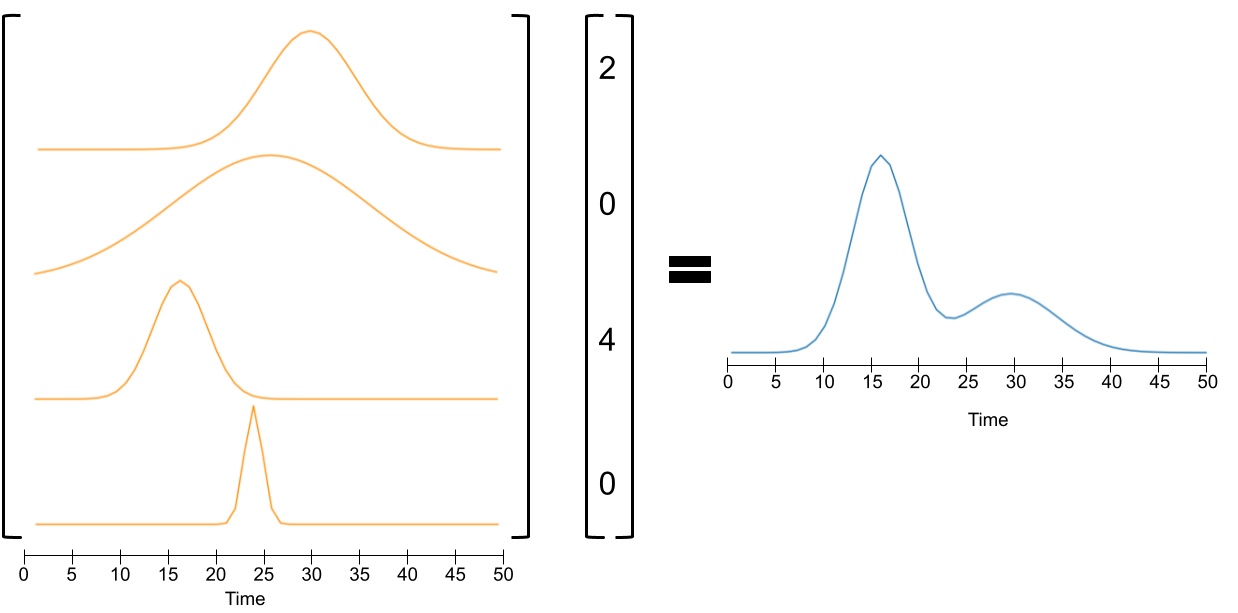}
}
\caption{The blue signal on the right is reconstructed as a linear combination of the orange signals in the dictionary on the right.}
\label{fig:Dictionary}
\end{figure}

We first propose a 
{\em dictionary-based approach},
that uses a predefined set of plausible curves,
that each describes the number of infected people over time; see Figure~\ref{fig:Dictionary}. 
It then finds a linear combination of a small 
(learned) subset of these curves 
that best explain the observed data. Each of the included curves (those with a non-zero coefficient in this linear combination) 
can be interpreted as an independent sub-population. 
We explored two types of prior curves: either
each is a (a)~normalized Gaussian probability density functions, or is an 
(b)~SIR curve with its own learned parameters and initial states.

The second method is mixture of 
(a fixed number, \M, of) 
fittable curves, where $M$, the number of sub-populations, is determined by the user.
We tested a mixture of $M=3$ Gaussian curves, and a mixture of $M=3$ SIR models. 
For the mixture of SIR models, 
we modified the classical SIR model to allow for shifts of the starting time-point of each SIR trajectory via a Kronecker delta function.
We evaluate these models based on both fit to the data, and also the accuracy
(mean absolute percentage error: MAPE),  for predicting the number of infected people 1-- to 4-weeks in advance.
Note this approach may help us to capture
the general trend of how the number of infected people change over time, and 
so 
might be very useful when the objective is to describe the main behaviour of the data.

\begin{figure}

\centerline{
 \includegraphics[width=\textwidth]{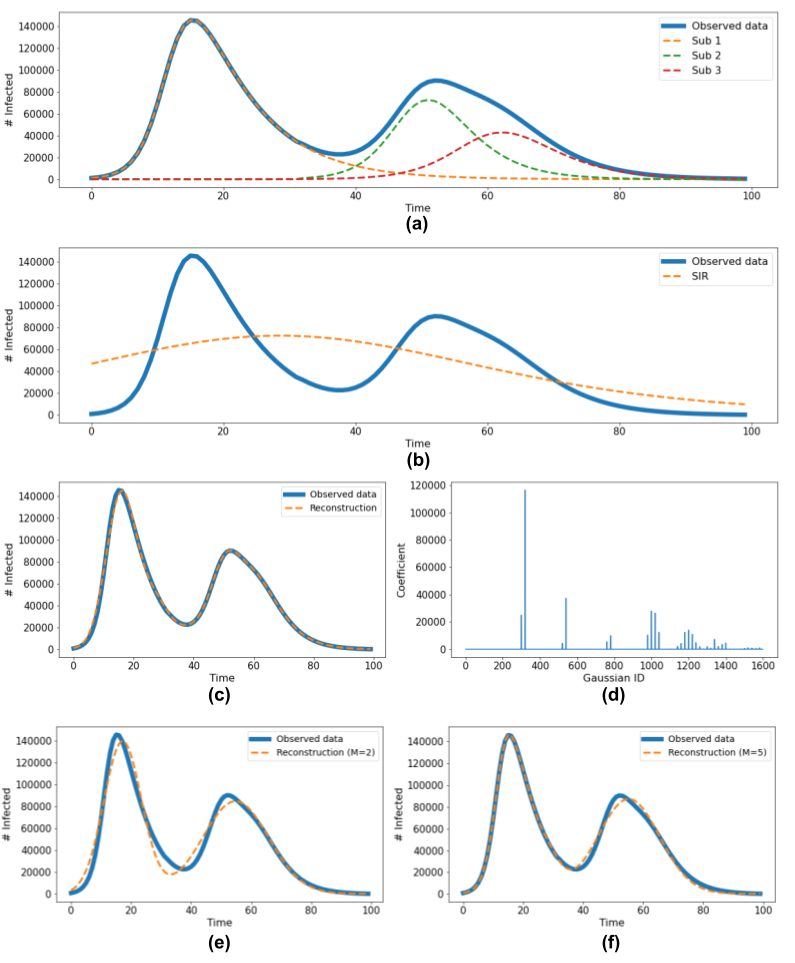}
}
\caption{Simple example showing the rationale behind our proposed approach. 
(a)~Simulated data consisting of three sub-populations, and the observed data.
(b)~Best fit of a classical (single) SIR model. 
(c)~Curve reconstructed by using Gaussian dictionaries; using (d)~the coefficients associated with each curve in the Gaussian dictionary. 
(e)~Curve reconstructed by the mixture of \M\ = 2 Gaussians. 
(f)~Curve reconstructed by the mixture of 
\M\ = 5 Gaussians.}
\label{fig:Simple_Example}
\end{figure}

To motivate 
our approach,
consider the three simulated 
independent ``SIR sub-populations"
shown by the dashed lines in  Figure~\hbox{\ref{fig:Simple_Example}}(a),
whose 
orange line represents the sub-population in a neighborhood A,
while the other two sub-populations 
are both located in a neighborhood B,
but 
have different behaviours -- green being more socially active than red.
But as the number of infected people are reported only at the state level
(not the neighborhood level),
we only observe the sum of the three sub-populations -- the solid blue line. 
This is the only information used by the algorithms modeling the number of infected people -- note they do not know have access to the number of sub-populations in the data,
nor the number of infections within each.

Figure~\ref{fig:Simple_Example}(b) shows the curve corresponding to the best fit of a classical SIR model. Note that it does a poor job describing the data, 
because the SIR model (and most of its extensions) are unimodal,
meaning they cannot accurately fit
multi-modal data. 

Our proposed approach models the observed data as a combination of several sub-populations. 
Since the number of sub-populations is unknown, the dictionary approach uses as many sub-populations as needed to create a curve that closely matches the observations (in terms of mean squared error). Figure~\ref{fig:Simple_Example}(c) shows how a Gaussian dictionary appropriately models the data, 
albeit using 114 Gaussian curves, each with the coefficient shown in Figure~\ref{fig:Simple_Example}(d). 
The mismatch between the number of real sub-populations and the number of sub-populations estimated by the model arises in part because the observed data is in fact not a mixture of Gaussians. 
In general, dictionary approaches can accurately model the observed data, provided they have a sufficiently large and diverse set of possible curves. However, this may require a large number of curves, which can make the resulting combination difficult to interpret and also can mean the forecasting version will overfit.

Figures~\ref{fig:Simple_Example}(e) and (f) show the curves reconstructed by the mixture of $M=2$ and $M=5$ fittable Gaussians. 
While the mixtures do capture the main trend of the data, 
they clearly have a higher error than the dictionary approach. 
Of course, as the number of curves, $M$, increases, the error decreases. 
Despite showing a higher error, 
as this approach lets the user control the number of curves to be used for modeling the data, 
it potentially 
allows an easier analysis over the identity of the curves. 
For example, it would be easier to 
interpret 
Figure~\ref{fig:Simple_Example}(e)
to mean that one curve belongs to a neighborhood A and the other to neighborhood B, 
than trying to do a similar analysis for the 114 curves selected from the Gaussian dictionary (Figure~\ref{fig:Simple_Example}(d) ).

\subsection{Related work} 
\label{sec:relatedWork}
Most of the efforts in modeling the evolution of epidemics with heterogeneous populations fall under the umbrella term 
\emph{meta-population models}~\citep{calvetti2020metapopulation,tan2021stochastic},
which  
divide the individuals into a specified number of sub-populations, each representing either a spatial area (a country, a city, a neighbourhood, etc.), or a known variable of interest (age, sex, etc).
They then use a compartmental model, such as an SIR or one of its extensions, for each sub-population, and perhaps also include some additional terms that describe the dynamics among the sub-populations~\citep{calvetti2020metapopulation,contreras2020multi,gerasimov2021covid,duan2022heterogeneous}. The sub-populations are usually allowed to have different interaction rates~\citep{albi2021control}. These models implicitly assume that the infections start at the same time across the different sub-populations. 
The quality of the overall model is tied to the quality of the selected sub-populations, 
and to the accuracy in the modeling of the interaction among them~\citep{kong2022compartmental}. Our work extends those results by allowing \hbox{\em latent}\ sub-populations, which can start at different times and can be of any parametric family.

There are some studies that use latent variables to model the uncertainty related to the reporting of newly infected people over time~\citep{bartolucci2022spatio,shi2022pan}.
Here, the latent variables represent the real number of new infections,
while the reported quantities are the observations associated with such latent variables, perhaps modeled using 
Hidden Markov Models. 
Some studies additionally incorporate the possibility of having observed sub-populations at the geographical level to account for interactions between populations~\citep{bartolucci2022spatio}. 
Similarly, \cite{tan2021stochastic} described a state-space model where latent variables represent the compartments of an 
SEIRD (Susceptible, Exposed, Infected, Recovered, Deceased) model, 
while the observations are only the confirmed new infections and cumulative number of deaths. 
They used a meta-population framework that models the spread of the disease across separate geographical sub-populations 
(\eg two different cities). 
Each sub-population might have different transition parameters, but their estimation requires the individual observations for each sub-population --
\ie the daily number of infections in city 1 and in city 2.
Of course, this is a different type of latency,
as this is still assuming a single homogeneous population, but where the values for some compartments are not observed.
By contrast, 
we propose the use of latent variables that represent the different sub-populations. 
Therefore, the observed data is a single time-series that is the \emph{sum}\ of the time-series of the individual sub-populations.

A different approach for modeling heterogeneous populations are the 
{\em agent-based models},
which model a sample of the individuals
in the population. 
\cite{nepomuceno2016individual} describe 
the general theoretical framework of an agent-based modeling approach for infectious disease dynamics, where each agent is an individual (potentially with different features based on their age, location, activities, etc.),
and the aggregation of individual behaviors is an estimate of the population dynamics as a whole. 
Meta-population models can be adjusted to approximate the dynamics of complex agent-based models~\citep{zachreson2022effects}. However, the usability of agent-based models is limited due to their computational complexity. 

Very relevant to our approach is the work of~\cite{sarker2022unifying}, 
who also consider latent sub-populations,
where the learner can only observe a single time-series
that is the aggregate of the different sub-populations. 
They model the number of infected people over time as a mixture of Gaussians, and propose using a peak-finding heuristic to determine the number of sub-populations (each modeled as a Gaussian) to use in their mixture. 
Our work extends theirs as we additionally show how to model an epidemic using a mixture of SIRs, 
and propose a general dictionary approach that allows us to model the number of infected people as a combination of any parametric model.

\section{Materials and Methods} 
\label{sec:Methods}
We compare three different methods to model the dynamics of the number of people infected with COVID-19: 
(1)~The traditional SIR model with 
learned parameters.
(2)~A dictionary approach that contains a set of predefined `candidate' curves that 
can be combined to describe the COVID-19 dynamics
(where each curve describes a sub-population) and an optimization algorithm determines which sub-populations to include (and with what weight).
(3)~A mixture approach that combines a predetermined number of parameterized curves
(\eg 2 or 3),
which uses an optimization algorithm to determine the parameters for each curve.

Importantly, methods~2 and 3 assume that the sub-populations are latent --
\ie we only observe the total number of infected people in the entire population, and not the number of infected people in each sub-population. 
Also, the `identity' of the sub-populations is a hidden variable --
\ie we do not know which people are included in each sub-population. 
This allow us to model sub-populations that are not even based on observed variables, such as sex, age and geographical location~\citep{kong2022compartmental}. 
This means each sub-population could perhaps represent people with some unobserved behaviour (\eg about vaccination, or masking) or some other set of characteristics.

\subsection{Traditional SIR model} 
\label{sec:SIR}
The SIR (Susceptible - Infected - Removed) model is a mathematical model commonly used to describe the dynamics of epidemics~\citep{kermack1927contribution}. 
It divides a population of $N$ individuals into 3 disjoint groups: \textbf{(S) Susceptible} = 
the people who have not been infected, but can potentially get the disease; 
\textbf{(I) Infected} = 
the people who are able to transmit the virus to the susceptible people;
\textbf{(R) Removed} = 
the people who recovered from the infection, or who died. 
This model assumes that the population size is constant ($N = S + I +R$) and is described by the set of differential equations:

\begin{equation}\label{eq:SIR}
    \begin{gathered}
      \frac{dS}{dt}\ =\ - \frac{\beta\, S(t)\, I(t) }{N}, \qquad
      \frac{dI}{dt}\ =\ \frac{\beta\, S(t)\, I(t) }{N}\ -\ \gamma \,I(t), \qquad 
      \frac{dR}{dt}\ =\ \gamma\, I(t)
    \end{gathered}
\end{equation}

\noindent where $S(t), I(t)$ and $R(t)$ represents the number of people in the susceptible, infected and removed compartments, respectively, at time $t$. $\beta$ represents the transmission rate (how quickly people move from susceptible to infected), and $\gamma$ represents the removed rate (how quickly infected people recover or die).

Following \cite{vega2022simlr}, we approximate
the differential equations of the SIR model with their discrete counterpart: 

\long\def\comment#1{}
\begin{equation}\label{eq:SIR_discrete}
  \begin{array}{llrr}
        S_t\ \ =& S_{t-1} &-\ \displaystyle \beta \frac{S_{t-1}\, I_{t-1} }{N}\ &\\[1.3ex]
        I_t \ \ =&I_{t-1}&+\ \displaystyle \beta\frac{ S_{t-1}\, I_{t-1} }{N}& -\ \gamma\, I_{t-1} \\
        R_t\ \ =&R_{t-1}&&+\ \gamma\, I_{t-1}
        \end{array}
\end{equation}

and estimate the optimal parameters $\beta^*$ and $\gamma^*$ by solving the optimization problem of Equation~\ref{eq:beta_gamma}:

\begin{equation}\label{eq:beta_gamma}
    \begin{gathered}
    (\beta^*,\ \gamma^*)\quad =\quad \argmin_{\beta,\gamma} \sum_{t=1}^n \left(I_t -  \left(\beta\,\frac{ S_{t-1} I_{t-1} }{N} \ - \ \gamma\, I_{t-1}\ + \ I_{t-1}\right)\right)^2
    \end{gathered}
\end{equation}

\noindent where $S_t, I_t, R_t$ are the number of individuals in the susceptible, infected and removed compartments, respectively, at time $t$;
$S_{t-1}, I_{t-1}, R_{t-1}$ represent the number individuals in each compartment at time $t-1$, $N\ =\ S_t + I_t + R_t$ is the time-invariant total number of people,
and $n$ represents the total number of points in the time-series used for training. In our experiments every time-step spans a week. 

Note that Equation~\ref{eq:beta_gamma} minimizes the error between the observations and predictions over just the number of people infected, 
but not over the numbers of susceptible or removed. This is because our purpose is to make predictions over the infections, whose number is several orders of magnitude smaller than the number of susceptibles.



\subsection{Dictionary-based models} 
A dictionary, $D \in \mathbb{R}^{S \times T}$, is a matrix that contains $S$ time-series, each with $T$ time-points. We can then reconstruct an observed time-series (\eg the number of people infected of COVID-19) as a linear combination of the elements in the dictionary:
\begin{equation}\label{eq:Dictionary_reconstruction}
    x = D^T \theta
\end{equation}
\noindent where $\theta \in \mathbb{R}^S$ contains the coefficient of each element in the dictionary. Figure~\ref{fig:Dictionary} depicts a simple example with a dictionary, $D$, that contains 4 time-series (shown in orange, on the left). 
The vector of weights
$\theta = 
\begin{bmatrix}2 & 0 & 4 & 0 \end{bmatrix}^T$ 
indicates the coefficients of the linear combination of the signals in the dictionary used to reconstruct the signal $x$ (shown in blue, on the right).

\def\nisp#1#2{I^{#1}_{#2}} 

We can think of the entries in the dictionary as candidate trajectories describing the number of people infected over time for a specific sub-population. These entries are computed 
{\em a priori}\ 
and should include plausible trajectories that we might expect to observe. 
There are different ways of computing these curves. For example, we can create a set of Gaussian probability density functions with different parameters for the mean, variance. Alternatively, we can simulate a set of SIR models with different parameters (initial number of people in each compartment at a specified time, and parameters $\beta$ and $\gamma$), and then include in the dictionary the curve corresponding to the number of infected people for that specific sub-population,
 $\nisp{r}{} = [\nisp{r}{0},\, \nisp{r}{1},\, \dots, \, \nisp{r}{T}]$%
.

We can then estimate the total number of infected people at each time
 $I = [I_0,\, I_1,\, \dots, \, I_k]$
 as a linear combination of the entries in the dictionary 
 $(\nisp{}{t}\ =\ \sum_{r=1}^M \theta_r \,\nisp{r}{t})$%
 , 
 which we can compare with the observed values. 
 The vector of weights $\theta$ determines which curves should be included in the linear combination (when $\theta_r \neq 0)$, as well as their coefficients.

Given an observed time-series, $x$, and a dictionary with candidate trajectories, $D$, we can estimate the parameter $\theta$ of Equation~\ref{eq:Dictionary_reconstruction} by solving the optimization problem:

\begin{equation}\label{eq:Optimization_dictionary}
\begin{gathered}
    \theta^*\quad =\quad \argmin_{\theta} ||x\, -\, D^T \theta ||^2_2\ \ +\ \ \lambda\, ||\theta ||^2_2\\
    \mbox{s.t.}\ \qquad 
    \theta_i \geq 0
\end{gathered}
\end{equation}

\noindent where $\lambda$ is a regularization parameter used to decrease the chance of overfitting. Note that Equation~\ref{eq:Optimization_dictionary} requires \emph{non-negative} coefficients for the candidate curves in the dictionary, as the entries in the dictionaries represent plausible trajectories over the number of infected people for different sub-populations, which can only be added, but not subtracted --
\ie the number of people infected in a sub-population cannot be negative.

Equation~\ref{eq:Optimization_dictionary} is a standard convex optimization problem with linear constraints. The indexes of the non-zero coefficients of the parameter $\theta$ represent the indexes of the sub-populations included in the model. 
Note that the number of non-zero indexes is automatically determined by the algorithm, so there is no need to manually identify the number of sub-populations to be included.

For our experiments, we compared the performance of two different dictionaries: 
(1)~$D_G$, whose entries are normalized probability density functions (PDF) of a Gaussian. 
We normalize each curve by dividing the entire PDF by its maximum.
The dictionary of Gaussians, $D_G$, contains 390 trajectories, obtained by using all possible combinations of a set of means $\mu \in \{0, 2, 4, \dots, 52\}$ and a set of variances $\sigma^2 \in \{1^2, 3^2, 5^2, \dots, 29^2\}$.
Intuitively, the mean $\mu$ represents the week at which we expect the peak of the number of infected people in a sub-population, 
while the standard deviation $\sigma$ represents the speed at which the number of infected people increases and decreases. 
If required, we could make the Gaussian distributions non-symmetric by controlling the skewness of the distribution~\citep{azzalini1999statistical}.

\def\DSIR{D_{SIR}}  
(2)~We also consider the $\DSIR$ dictionary, whose entries are each generated by taking the number of people in the infected compartment of a SIR model initialized with different parameters.
For each, we need to specify the parameters $\beta$ and $\gamma$ of the traditional SIR model, and also the initial number of susceptible, infected and removed at the beginning of the simulation. We also need to create time-shifts in the traditional SIR model --
\emph{\eg} to specify that the number of infected people will 
start increasing after 10 weeks --
\ie become non-zero after week~10.

To do this, our model specifies that the initial number of people in the infected and recovered compartments is zero, 
and then modifies the discrete-time SIR model of Equation~\ref{eq:SIR_discrete} as follows:
\begin{equation}\label{eq:SIR_Kronecker}
    \begin{array}{llrrr}
    S_{t}\ &=\ S_{t-1}&-\ \ S_{t-1}\, I_{t-1}\, \beta 
                  & -\ C\ \delta(t - k)\ \\
    I_{t}\ &=\ I_{t-1}&+\ \ S_{t-1}\, I_{t-1}\, \beta
                & +\ C\ \delta(t - k)\ &-\ \gamma\, I_{t-1}\\
    R_{t}\ &=\ R_{t-1}& && +\ \ \gamma\, I_{t-1} 
    \end{array}%
\end{equation}

\noindent where $\delta(\cdot)$ is the Kronecker delta function,
$k$ is the `start time' for this sub-population, 
and $C$ 
is a 
parameter
that 
specifies how many people will be incorporated into this infected compartment at time $k$.

Using this model of Equation~\ref{eq:SIR_Kronecker},
we created a dictionary $D_{SIR}$ with 546 curves obtained by using all possible combinations of the parameters $S_0 \in \{10^4, 10^5, 10^6\},\ 
\beta \in \{0.3, 0.4, \dots 0.9\},\ 
k\in \{ 0, 2, \dots, 50\}$,
where we also set 
$\gamma = 0.5,\ I_0=0,\ R_0=0,\ 
C = 100$. 
Similarly to the dictionary with Gaussians, we normalized the curves of the infected population by dividing over the maximum value of each curve, so the peak of each curve is always one.

\def\apriori{{\em a priori}\ }
\subsection{Fixed size mixture of fittable curves model}

The dictionary-based models have the advantage of automatically finding the number of sub-populations to include in the model, as well as the coefficient associated with each sub-population; 
however, it might be useful to instead just specify 
\apriori\ 
the number, $M$, of sub-populations that we want to include but here allow the system to fit the associated parameters. Controlling the number of non-zero coefficients in 
Equation~\ref{eq:Optimization_dictionary} requires a 0-norm regularization term, which is intractable for large models 
(\hbox{\eg} when there are a large number of curves in a dictionary).

Instead, we can use different parametric models to fit the observed data. For example, given a sequence of T time-points, $x = [0, 1, \dots, T]$:
\begin{equation}
    f(\,x,\, \theta,\, \mu, \,\sigma\,)
    \ \ =\ \ \theta_0\ +\ \sum_{m=1}^M \theta_m \ \exp \left(-\frac{(x-\mu_m)^2}{2\sigma_m^2} \right)
\end{equation}

Then, given an observation of the number of infected people over time, $I = [I_1, I_2, \dots, I_T]$, we find the parameters of the Gaussian mixture that solve the optimization problem:
\begin{equation}\label{eq:mixture_Gaussians_optimization}
    \theta^*, \mu^*, \sigma^*\quad =\quad \argmin_{\theta, \mu, \sigma} || y\ -\  f(x, \theta, \mu, \sigma)||_2^2
\end{equation}

Alternatively, we can model the number of infected people by taking the Infected compartment of a mixture-of-\M\ fittable SIR models. 
Each SIR model in the mixture follows the set of Equations~\ref{eq:SIR_Kronecker}. 
Therefore, the number of people infected at time $t$ is given by:
\begin{equation}\label{eq:mixture_SIR}
    f(S_{t-1}, I_{t-1}, \beta, \gamma, C, k)\ =\  \sum_{m=1}^M S_{t-1}^m I_{t-1}^m\beta^m +C^m \delta(t-k^m) - \gamma^m I_{t-1}^m + I_{t-1}^m
\end{equation}

There are two important things to note in Equation~\ref{eq:mixture_SIR}. First, each of the terms $S_{t-1}, I_{t-1}, \beta, \gamma, C$ and $k$ is a vector of length $M$, 
whose $i$-th entry represents the value of that parameter for the 
$i$-th sub-population. 
Second, we do not observe the values of any of these vectors,
but instead 
only observe the aggregate quantity $I_t\ =\ \sum_m I_{t,m}$.
That is, 
we observe the total reported number of infected people, but we do not know how many infected are in each sub-population.

Since Equation~\ref{eq:mixture_SIR} is recursive, we can compute a time-series of length T if we have the values 
$S_{t=0},\ I_{t=0},\ \beta,\ \gamma,\ C$ and $k$. 
Given a sequence of T observations $I = [I_1, I_2, \dots, I_T]$ that represent the number of reported infections over time, 
we assume that $I_{t=0} = [0]^M$ and 
we can find the 
remaining 
parameters by solving the optimization problem:
\begin{equation}\label{eq:mixture_SIR_optimization}
    S_0^*, \beta^*, \gamma^*, C^*, k^*\ 
    =\ \argmin_{S_0, \beta, \gamma, C, k} ||y - \hat{y}(S_0, \beta, \gamma, C, k, T) ||_2^2
\end{equation}
\noindent where $\hat{y}(S_0, \beta, \gamma, C, k)$ is a time-series of length T obtained by recursively applying Equation~\ref{eq:mixture_SIR} $T$ times.

Although the optimization problems of Equations~\ref{eq:mixture_Gaussians_optimization} and~\ref{eq:mixture_SIR_optimization} are not convex, 
we can find a locally optimal solution using optimization techniques such as generalized simulated annealing~\citep{xiang2000efficiency}. 
This optimization technique requires specifying a tuple (lower bound$_j$, upper bound$_j$) that sets the range of possible valid values for the $j^{th}$ variable to be optimized. 
For optimizing 
Equation~\ref{eq:mixture_Gaussians_optimization}, 
we used $\mu \in (0,50), \sigma \in (1,6)$ and $\theta \in (0,3 \times 10^5)$,
and for optimizing Equation~\ref{eq:mixture_SIR_optimization},
we used $S \in (0,10^8)$, $\beta \in (0,1)$, $\gamma \in (0,1)$, $C \in (0,10^3)$, and $k \in (0,50)$. For both equations we set $M=3$ to illustrate the performance of the algorithm. 
In a real application, the number of curves will change for every geographical location; however, note that the number of parameters in the search space increases exponentially with $M$. 
Therefore, we only recommend this method 
for small values of $M$.
For larger values, the dictionary approach is more efficient.

\begin{figure}
\centerline{
 \includegraphics[width=0.7\textwidth]{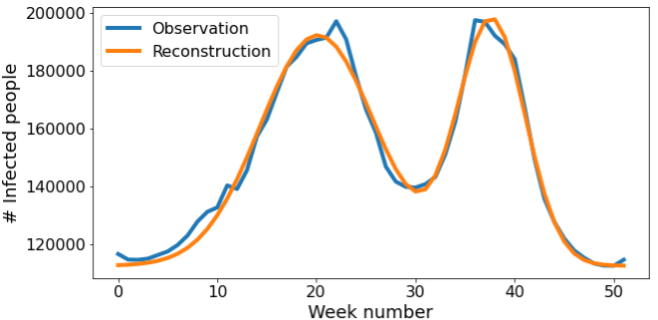}
}
\caption{Reconstruction of the number of weekly number of people infected in Canada during 52 weeks, 
from 30 July 2020 to 29 July 2021. 
The blue line represents the observed data, while the orange line represents the reconstruction obtained by using a mixture of 2 Gaussians.}
\label{fig:M_Gaussians}
\end{figure}

\section{Results}
We evaluated the performance of the methods described in Section~\ref{sec:Methods}, 
in two different tasks: 
(T1)~Accurate modelling of the number of infected people, and (T2)~Forecasting of the number of infected people 1 to 4 weeks in advance. For T1 (modelling the pandemic) we use the same data 
for training and evaluation, 
but T2 (Forecasting)
requires a training set for learning the parameters, and a disjoint set to assess the performance. 
Here, for each $t$,
we predicted the number at time $t+1$ based on the data observed from time $0, \dots, t$.
For both tasks,
we evaluated the learned model using the mean absolute percentage error (MAPE):
\begin{equation}
    \mbox{MAPE}\ \ =\ \ \frac{1}{n}\ \sum_{t=1}^n \left| \frac{A_t - F_t}{A_t}\right|
\end{equation}
\noindent where $n$ is the total number of time-points, $A_t$ is the actual value observed at time $t$, and $F_t$ is the forecasted value at time $t$.

For both tasks, we used the publicly available COVID-19 Data Repository by the Center for Systems Science and Engineering at Johns Hopkins University~\citep{dong2020interactive}, and followed the pre-processing pipeline suggested by~\cite{vega2022simlr}. We restricted our analysis to the time period between 
30 July 2020 and 29 July 2021
-- selecting this 
time frame because it includes two or more ``COVID waves", and also because 187 countries were consistently reporting the number of new people infected during that period. 

\subsection{Modeling task (T1)}
\begin{figure}
\centerline{
 \includegraphics[width=0.8\textwidth]{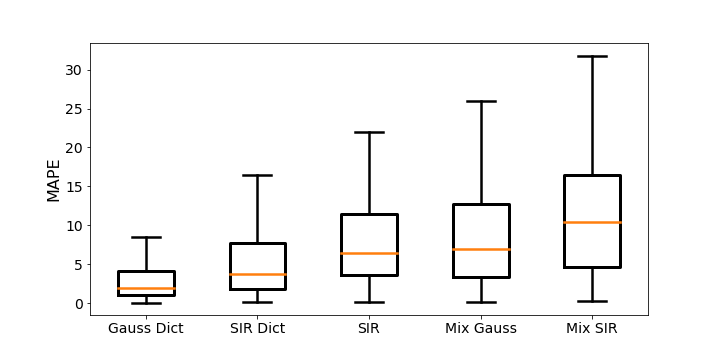}
}
\caption{Visualization of the distributions of the modeling task, in terms of mean absolute percentage error (MAPE), for the different modeling methods.}
\label{fig:boxplot_reconstruction}
\end{figure}

Figure~\ref{fig:boxplot_reconstruction} shows the distribution of the MAPE over all 187 countries,
for the modeling task, for 
each of the models considered:
the classical SIR model, the dictionary models based on Gaussian distribution (resp., SIRs), as well as the mixture models of Gaussian distributions (resp., SIRs). Table~\ref{tab:reconstruction} provides the corresponding numerical values.

\begin{table}
  \caption{MAPE statistics of the different methods in the modeling task.}
  \label{tab:reconstruction}
  \centering
  \begin{tblr}{
      colspec={lccccccc},
      row{1}={font=\bfseries},
      column{1}={font=\itshape},
      row{even}={bg=gray!50},
    }
              & Mean  & Std  & Min  & 25\% & 50\% & 75\% & Max  \\
    \toprule
    Gauss Dict & 3.5 & 4.7 & 0.1 & 1.1 & 1.9 & 4.1 & 38.5 \\
    SIR Dict & 7.5 & 13.3 & 0.1 & 1.8 & 3.8 & 7.7 & 108.9 \\
    SIR & 8.7 & 8.2 & 0.1 & 3.6 & 6.5 & 11.5 & 64.5 \\
    Mix Gauss & 11.3 & 13.9 & 0.2 & 3.3 & 7.0 & 12.7 & 76.0 \\
    Mix SIR & 13.7 & 16.3 & 0.3 & 4.6 & 10.5 & 16.5 & 121.2 \\
    \bottomrule
  \end{tblr}
\end{table}

To compute the MAPE, we use the entire time-series as an input for each of the methods and learn the parameters that optimize Equations~\ref{eq:beta_gamma},~\ref{eq:Optimization_dictionary},~\ref{eq:mixture_Gaussians_optimization}, or ~\ref{eq:mixture_SIR_optimization}, depending on the method selected. Then, we used the time-series $x_0, x_1, \dots, x_t$, along with the learned parameters, to predict $x_{t+1}$ for $t = 1, \dots, 52$. We applied this procedure individually to each of the 187 countries in the dataset.

Figure~\ref{fig:boxplot_reconstruction} and Table~\ref{tab:reconstruction} show that the dictionary based approaches not only have the lowest median MAPE, but also the smallest variance. Both dictionary based approaches show a better performance than the standard SIR model. Finally, the mixtures of M fittable curves (for $M=3$) have a higher median error and higher variance than traditional SIR models.

\begin{figure}
\centerline{
 \includegraphics[width=0.9\textwidth]{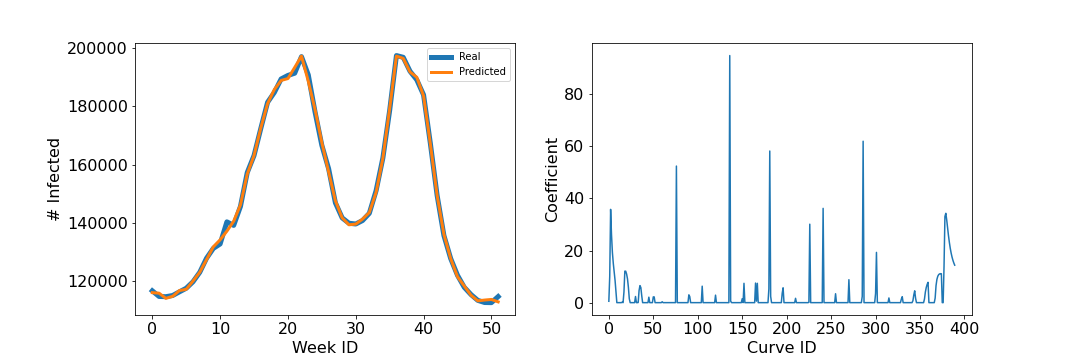}
}
\caption{Visualization of the weights learned by the Gaussian dictionary approach in data from Canada.}
\label{fig:weight_visualization}
\end{figure}

While both the dictionary-based methods and the mixture of \M\ fittable curves compute mixtures of curves, they differ in that dictionary-based approaches automatically decide on the number of curves to use, as well as their respective coefficients. For example, Figure~\ref{fig:weight_visualization} shows the coefficients learned by the algorithm based on Gaussian dictionaries; the non-zero coefficients, which here is 32, are the number of selected sub-populations. 
By contrast, the mixture of \M\ fittable curves approach requires the user to explicitly set the number of mixtures to use. 

\begin{figure}
\centerline{
 \includegraphics[width=0.9\textwidth]{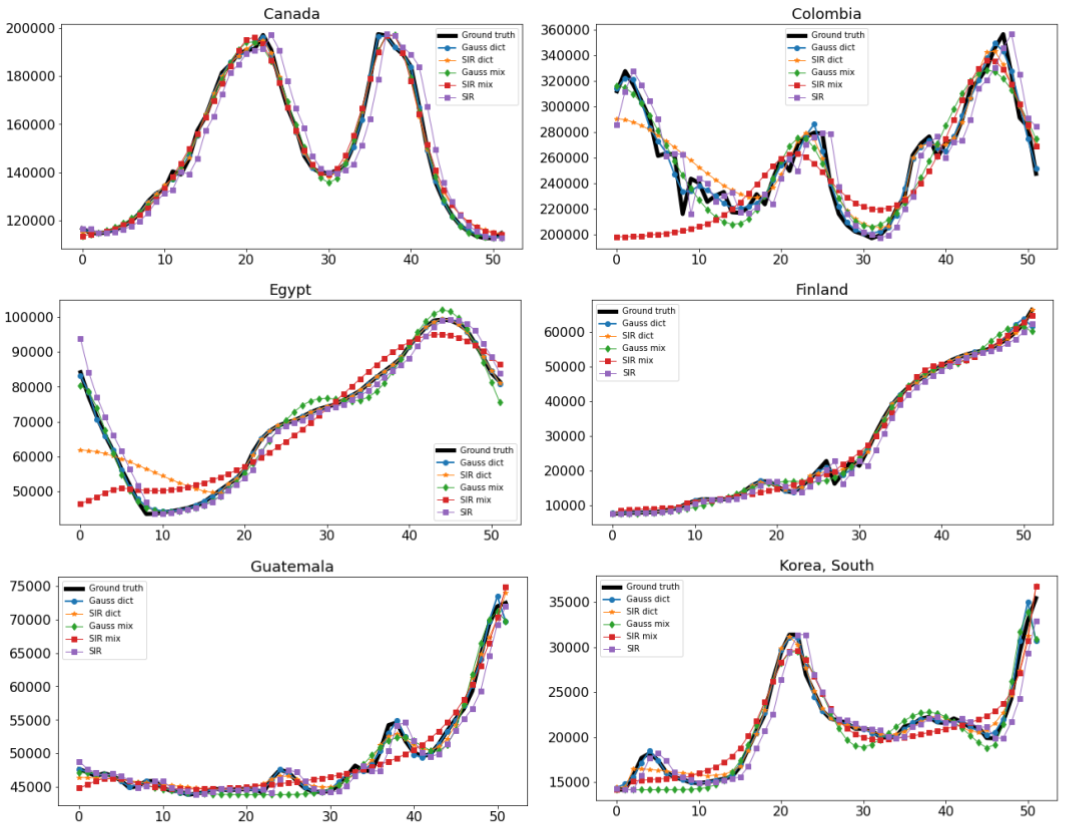}
}
\caption{Examples of the resulting graphs produced by the different models in the reconstruction task. The plots show the weekly number of infected people in different countries.}
\label{fig:examples_reconstruction}
\end{figure}

Figure~\ref{fig:examples_reconstruction} depicts examples of curves generated, for modeling the number of infected people over time, by each of the 5 methods, for 6 different countries. We see that the Gaussian dictionary approach is able to accurately model trajectories that are multimodal --
\ie it can properly model different waves. 
It also can smooth trajectories that have high-frequency oscillations
-- \eg see Colombia. This smoothing effect is more visible for the dictionaries based on SIR (see the behaviour of the first 15 weeks in Colombia, Egypt, and South Korea).

While the traditional SIR model performs fairly well for uni-modal trajectories – like Egypt, Finland, and Guatemala – it struggles with multimodal time-series, such as Canada, Colombia and South Korea. When there multiple waves, the model basically produces (an almost) delayed version of the original data. This occurs because the SIR is unimodal by definition, 
so most of the time the prediction for $x_{t+1}$ is the value very close to the value observed at $x_t$.

Finally, the mixture of \M\ fittable curves have a higher error, on average, than the other approaches. While these models do capture the general trend of the data, as they have a limited number of curves (3 in this case), they are unlikely to capture the minor variations of the original data. Also, note that the mixture of SIR models always start at the minimum possible value, so it fails to properly model cases where the minimum number of infected people is not at the beginning of the training sequence (see the cases of Colombia and Egypt).

Despite their inferior performance, 
the mixtures of M fittable curves might be useful for understanding and explaining the dynamics of the number of infected people over time. 
For example, in the cases of Canada and Colombia in Figure~\ref{fig:examples_reconstruction}, it might be easier to explain the number of infected people in terms of 2 or 3 waves, respectively, than to give precise meaning to all the non-zero coefficients depicted in Figure~\ref{fig:weight_visualization}. 
Also, note that the ``ground truth'' is the reported number of infections over time, 
which in general is only an approximation of the actual number of infected people over time 
(which is in fact unknown). 
Therefore, in some situations it might be desirable to model the pandemic with a method that ignores the small changes in the time-series, which might be just noise.

\begin{figure}
\centerline{
 \includegraphics[width=\textwidth]{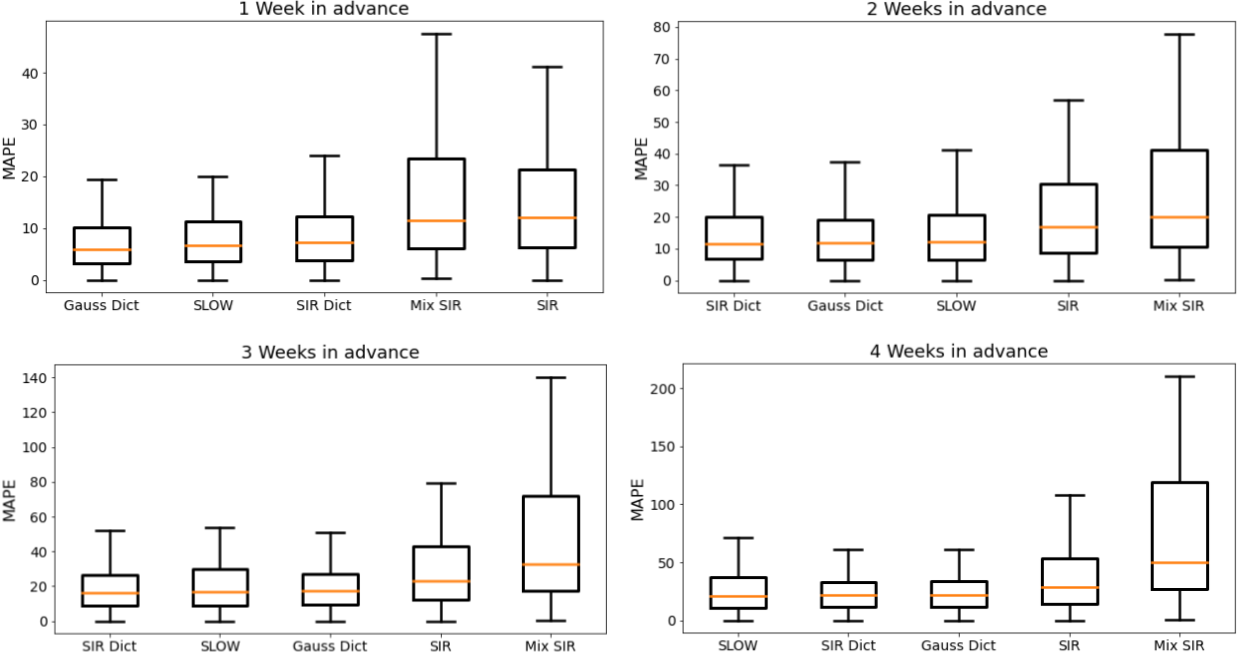}
}
\caption{Visualization of the distributions of the predictions 1-to-4 weeks in advance, in terms of mean absolute percentage error (MAPE), for the different predictions methods.}
\label{fig:forecasting_graph}
\end{figure}

\subsection{Forecasting task (T2)} 
Here, we compare 6 different approaches for forecasting the number of infected people 1- to 4-weeks in advance: 
the classical SIR model, the dictionary approaches based on Gaussian distributions and SIRs, the mixture of $M=3$ fittable SIR models and $M=3$ Gaussian models, 
and the ``same as last observed week'' method (SLOW) method. 
Figure~\ref{fig:examples_reconstruction} 
plots the distribution of the MAPEs for each approach for forecasting the number of infected people, 1- to 4-weeks in advance; this is over all 187 countries.
Their numerical values are shown in Tables~\ref{tab:week_1} to~\ref{tab:week_4}. 
As the values for the mixture of Gaussians are so much higher than the other methods, we did not include them in 
Figure~\ref{fig:forecasting_graph} for an easier visualization.

Unlike the modeling task, 
which learns the model's optimal parameters 
after observing the \emph{entire} dataset, 
in the forecasting task, each approach forecasts the number of infected people in time $t+1$, $x_{t+1}$ (and also $x_{t+2}, x_{t+3}$ and $x_{t+4})$ 
based on only the models learned from the observations $x_0, \dots, x_t$. 
We repeat this process iteratively from $t=5$ to $t=48$.

Here, we included SLOW as a baseline method. 
While this method is very naive,
\cite{vega2022simlr} showed that this approach did quite well -- indeed, better than many other approaches to this task; see the results of the COVID-19 Forecasting Hub~\citep{Cramer2021-hub-dataset}, which is the official forecasting project of the Center for Disease Control and Prevention (CDC) in the United States. 

Similar to the reconstruction task, the dictionary-based approaches outperform the classical SIR model in terms of MAPE.
Their variance is also the smallest across the different methods used in this comparison and
they are the only approaches that perform better than the baseline model (SLOW). 
The mixture of \M\ fittable curves, again, have the worst performance with the highest variance.

\begin{figure}
\centerline{
 \includegraphics[width=\textwidth]{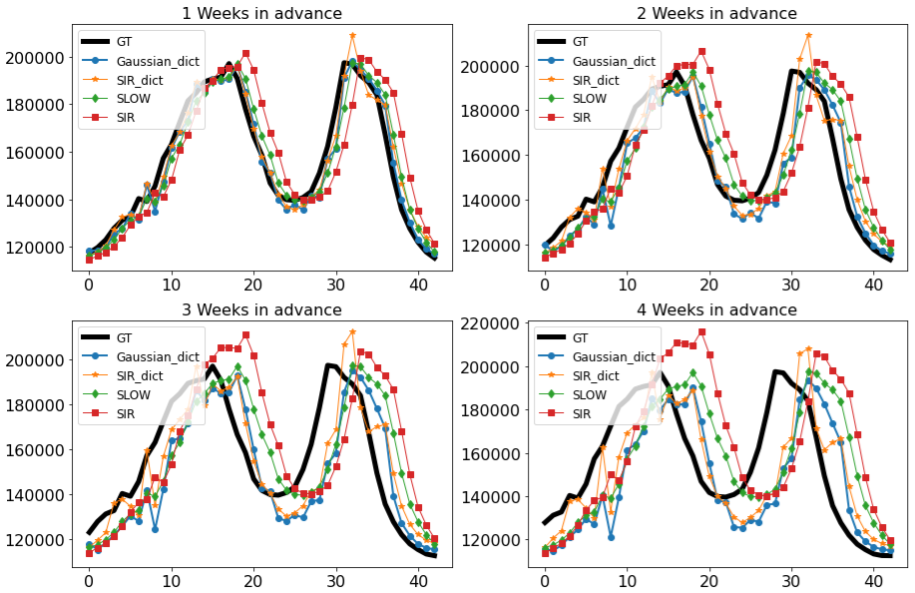}
}
\caption{Examples of the curves predicted by the different methods for Canada 1-to-4 weeks in advance. (GT = ground truth).}
\label{fig:examples_forecasting}
\end{figure}

As expected, the MAPE increases with longer forecasting horizons. 
While the median MAPE of the Gaussian dictionary is 6\% for 1-week forecasting, it increases to 11.8\%, 17.1\%, and 21.7\% 
when forecasting 2, 3, and 4 weeks ahead, respectively.
The other methods follow a similar behaviour. 
Figure~\ref{fig:examples_forecasting} shows an example of the predictions made by the different methods for the Canadian dataset. Note how the dictionary based approaches tend to make more accurate predictions than the SIR, SLOW, and mixture models.

\begin{table}
  \caption{Prediction results 1 week in advance}
  \label{tab:week_1}
  \centering
  \begin{tblr}{
      colspec={lccccccc},
      row{1}={font=\bfseries},
      column{1}={font=\itshape},
      row{even}={bg=gray!50},
    }
              & Mean  & Std  & Min  & 25\% & 50\% & 75\% & Max  \\
    \toprule
    Gauss Dict & 8.1 & 8.7 & 0.0 & 3.2 & 6.0 & 10.1 & 71.0 \\
    SIR Dict & 10.8 & 16.8 & 0.0 & 3.7 & 7.2 & 12.3 & 152.6 \\
    SIR & 15.9 & 16.2 & 0.0 & 6.3 & 12.0 & 21.2 & 128.5 \\
    Mix Gauss & 31.2 & 39.5 & 0.0 & 7.7 & 16.0 & 38.5 & 247.4 \\
    Mix SIR & 165.2 & 1350.7 & 0.2 & 6.1 & 11.7 & 23.8 & 18127.3 \\
    SLOW & 8.7 & 8.7 & 0.0 & 3.6 & 6.7 & 11.2 & 68.9 \\
    \bottomrule
  \end{tblr}
\end{table}

\begin{table}
  \caption{Prediction results 2 weeks in advance}
  \label{tab:week_2}
  \centering
  \begin{tblr}{
      colspec={lccccccc},
      row{1}={font=\bfseries},
      column{1}={font=\itshape},
      row{even}={bg=gray!50},
    }
              & Mean  & Std  & Min  & 25\% & 50\% & 75\% & Max  \\
    \toprule
    Gauss Dict & 14.8 & 14.5 & 0.0 & 6.5 & 11.8 & 19.1 & 128.5 \\
    SIR Dict & 16.5 & 21.7 & 0.0 & 6.6 & 11.5 & 20.3 & 174.9 \\
    SIR & 24.1 & 25.7 & 0.0 & 8.7 & 16.8 & 30.4 & 200.0 \\
    Mix Gauss & 228.1 & 492.2 & 0.1 & 22.2 & 54.9 & 173.8 & 3484.4 \\
    Mix SIR & 1E9 & 2E10 & 0.3 & 10.5 & 20.4 & 42.1 & 2E11 \\
    SLOW & 15.9 & 15.7 & 0.0 & 6.4 & 12.1 & 20.7 & 125.5\\
    \bottomrule
  \end{tblr}
\end{table}

\begin{table}
  \caption{Prediction results 3 weeks in advance}
  \label{tab:week_3}
  \centering
  \begin{tblr}{
      colspec={lccccccc},
      row{1}={font=\bfseries},
      column{1}={font=\itshape},
      row{even}={bg=gray!50},
    }
              & Mean  & Std  & Min  & 25\% & 50\% & 75\% & Max  \\
    \toprule
    Gauss Dict & 20.8 & 20.4 & 0.0 & 9.3 & 17.1 & 26.9 & 182.8 \\
    SIR Dict & 22.3 & 27.2 & 0.0 & 9.1 & 16.5 & 26.8 & 229.1 \\
    SIR & 33.2 & 36.1 & 0.0 & 12.1 & 22.8 & 43.1 & 270.9 \\
    Mix Gauss & 1222.1 & 3381.4 & 0.8 & 38.6 & 111.9 & 519.4 & 23244.3 \\
    Mix SIR & 4E23 & 6E24 & 0.3 & 18.1 & 33.7 & 72.1 & 8E25 \\
    SLOW & 22.5 & 22.7 & 0.0 & 8.7 & 16.9 & 29.6 & 185.8 \\
    \bottomrule
  \end{tblr}
\end{table}

\begin{table}
  \caption{Prediction results 4 weeks in advance}
  \label{tab:week_4}
  \centering
  \begin{tblr}{
      colspec={lccccccc},
      row{1}={font=\bfseries},
      column{1}={font=\itshape},
      row{even}={bg=gray!50},
    }
              & Mean  & Std  & Min  & 25\% & 50\% & 75\% & Max  \\
    \toprule
    Gauss Dict & 25.4 & 23.2 & 0.0 & 11.8 & 21.7 & 33.5 & 220.5 \\
    SIR Dict & 27.3 & 29.7 & 0.0 & 11.9 & 21.5 & 33.6 & 223.0 \\
    SIR & 44.7 & 56.8 & 0.0 & 14.6 & 28.4 & 53.6 & 461.3 \\
    Mix Gauss & 3506.1 & 10909.3 & 2.1 & 42.1 & 153.1 & 1016.6 & 71112.5 \\
    Mix SIR & 4E52 & 5E53 & 0.4 & 27.4 & 50.7 & 119.7 & 7E54 \\
    SLOW & 28.3 & 27.7 & 0.0 & 11.0 & 21.1 & 37.3 & 234.1 \\
    \bottomrule
  \end{tblr}
\end{table}

\section{Discussion} 
\label{sec:Discussion}
Despite their popularity, SIR models and their extensions are usually unimodal, which limits their usability in time-series with several infection waves. These models also assume an homogeneous population. These assumptions appear invalid in many real life scenarios. Here we presented a method to model heterogeneous populations as a linear combination of several independent sub-populations, where the details of these sub-populations
(\eg which person is in which sub-population) 
is unknown. 
This approach naturally allows us to model different waves of the pandemic. We offer two alternatives to set the number of subpopulations to include in the model:
to manually define the number of sub-populations (mixture of \M\ fittable curves model),
or to use an optimization algorithm that automatically determines how many sub-populations to use, among a set of candidate trajectories (Dictionary based approach).

Dictionary based approaches showed a better performance, in terms of median and variance of the MAPE, on both the modeling and the forecasting task; however their performance strongly depends on the curves used to create the dictionary. 
Also, as shown in Figure~\ref{fig:weight_visualization}, the optimization algorithm selects a relatively high number of curves. Although we could interpret each of these curves as a sub-population, it would be difficult to identify the exact nature of them. Of course, we never claim that any of our approaches are learning causality, so even when they accurately model the dynamics of the number of infected people over time, there is no way of determining if in fact those are the real number of sub-populations present in the data. Despite this limitation in interpretability, these models proved to be more accurate than the rest in the short-term forecasting task.

The mixture-of-$M$ curves, on the other hand, had the worst performance,
being even lower than traditional single SIR models. 
Nevertheless, the mixture-of-$M$ curves might prove useful when the objective is to describe the general trend of the number of infected people over time. 
Figure~\ref{fig:examples_reconstruction} 
shows that the mixture models ignore the 
high-frequency oscillations in the data, 
identifying only the ``main waves'' present in the observed time-series. 
Since users have control over the number of sub-populations to model, it might be easier to assign a meaning to the different sub-populations in the model.

There have been several earlier attempts to include sub-populations in the SIR models; 
however, in most of these attempts the infections in the different sub-populations ``start'' at the same time.
Here, we propose the use of the Kronecker delta in the SIR equations to allow for the modelling of a combination of SIR's that start at different points in time.

Importantly, the dictionaries and the mixture-of-$M$ fittable curves are general methods for building mixtures of models. 
We presented examples based on Gaussian distributions and SIR models, and showed their effectiveness over 187 countries; 
however, this is a general approach that is compatible with any other parametric model,
such as SEIR and other variants of the classical SIR model. 
We expect that this approach can be used to model not only COVID-19, but other infectious diseases as well. 
To facilitate the exploration of this approach, the code for reproducing these results and for implementing the different approaches here presented is publicly available at \url{https://github.com/rvegaml/SubpopulationModels}

\bibliographystyle{abbrvnat}
\bibliography{main}

\end{document}